\documentclass[conference]{IEEEtran}

\usepackage{caption}
\usepackage{subcaption}
\ifCLASSINFOpdf
  \usepackage[pdftex]{graphicx}
  \graphicspath{{./graph/}}
  \DeclareGraphicsExtensions{.pdf,.jpeg,.png}
\else
  \usepackage[dvips]{graphicx}
  \graphicspath{{./graph/}}
  \DeclareGraphicsExtensions{.eps}
\fi

\ifCLASSINFOpdf
\else
\fi

\usepackage[cmex10]{amsmath,mathtools}
\usepackage{mdwmath}
\usepackage{mdwtab}
\usepackage{amssymb}
\usepackage{xcolor}
\usepackage{algorithmic}
\usepackage{algorithm}
\usepackage{float}

\usepackage{epstopdf}

\usepackage{color}

\usepackage[utf8]{inputenc}

\usepackage{cite}
\usepackage{bm}
\usepackage{hyperref}

\begin{document}

\title{Undermining Federated Learning Accuracy in EdgeIoT via Variational Graph Auto-Encoders}

\author{\IEEEauthorblockN{Kai Li\IEEEauthorrefmark{1}\IEEEauthorrefmark{2},
Shuyan Hu\IEEEauthorrefmark{3},
Bochun Wu\IEEEauthorrefmark{3},
Sai Zou\IEEEauthorrefmark{4},
Wei Ni\IEEEauthorrefmark{5},
and Falko Dressler\IEEEauthorrefmark{1}}
\IEEEauthorblockA{\IEEEauthorrefmark{1}Telecommunication Networks (TKN), TU Berlin, Germany.\\ }
\IEEEauthorblockA{\IEEEauthorrefmark{2}CISTER Research Centre, Portugal.\\ 
Email: kaili@ieee.org \& dressler@ccs-labs.org.}
\IEEEauthorblockA{\IEEEauthorrefmark{3}Fudan University, China.\\
Email: \{syhu14,wubochun\}@fudan.edu.cn.}
\IEEEauthorblockA{\IEEEauthorrefmark{4}Guizhou University, China.\\
Email: dr-zousai@foxmail.com.}
\IEEEauthorblockA{\IEEEauthorrefmark{5}CSIRO, Australia.\\
Email: wei.ni@csiro.au.}
}

\maketitle

\IEEEcompsoctitleabstractindextext{%
\begin{abstract}
EdgeIoT represents an approach that brings together mobile edge computing with Internet of Things (IoT) devices, allowing for data processing close to the data source. Sending source data to a server is bandwidth-intensive and may compromise privacy. Instead, federated learning allows each device to upload a shared machine-learning model update with locally processed data. However, this technique, which depends on aggregating model updates from various IoT devices, is vulnerable to attacks from malicious entities that may inject harmful data into the learning process. This paper introduces a new attack method targeting federated learning in EdgeIoT, known as data-independent model manipulation attack. This attack does not rely on training data from the IoT devices but instead uses an adversarial variational graph auto-encoder (AV-GAE) to create malicious model updates by analyzing benign model updates intercepted during communication. AV-GAE identifies and exploits structural relationships between benign models and their training data features. By manipulating these structural correlations, the attack maximizes the training loss of the federated learning system, compromising its overall effectiveness.
\end{abstract}

\begin{keywords}
Internet of Things (IoT), mobile edge computing, federated learning, variational graph auto-encoder, manipulating model accuracy, adversarial attacks
\end{keywords}}

\maketitle

\IEEEdisplaynotcompsoctitleabstractindextext
\IEEEpeerreviewmaketitle

\section{Introduction}
\label{sec_intro}
Edge-based Internet of Things (EdgeIoT) represents a critical advancement that merges mobile edge computing with IoT devices, enabling data processing close to the data source~\cite{fotia2023trust}. By combining the computational power at the edge with the flexibility of IoT devices, this method enables real-time data analysis and decision-making, minimizing delay and ease bandwidth demands that arise from transmitting data to a server. It is particularly crucial for applications that need rapid responses, such as disaster management, environmental monitoring, and smart city development. EdgeIoT revolutionizes the way that data is processed and utilized, paving the way for efficient, agile, and intelligent systems in metaverse~\cite{li2022internet,wen2025adaptive,zhang2018continuous,jiao2024medusa3d,zhao2024rgbe,zhang2024swift}.

\begin{figure}[htb]
\centering
\includegraphics[width=3.4in]{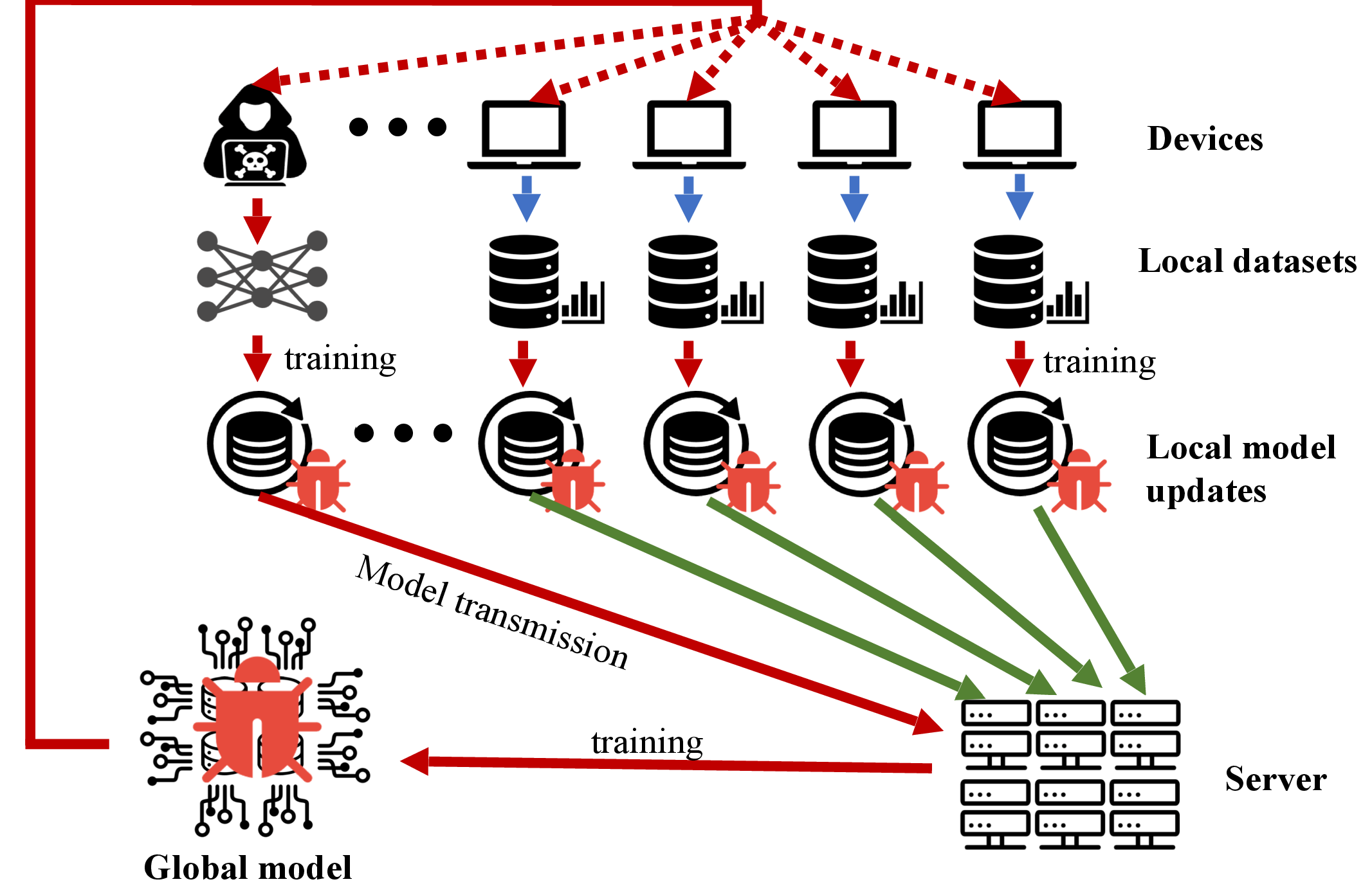}
\caption{Federated learning-enabled EdgeIoT involves IoT devices equipped with sensors and computational units that process data locally.}
\label{fig_application}
\end{figure}

Federated learning can greatly enhance data processing efficiency and decision-making in EdgeIoT environments. As illustrated in Fig.~\ref{fig_application}, federated learning within an EdgeIoT framework involves IoT devices equipped with sensors and computational units that process data locally~\cite{li2023towards}. Sending source data to a server is communication resources-intensive and may compromise privacy. Instead, federated learning allows each device to process data locally and only to upload the updated local model~\cite{zheng2022exploring}. However, the reliance on aggregating updates from multiple devices makes the system vulnerable to attacks. Malicious devices can inject harmful updates, which may distort the learning model and lead to incorrect or dangerous decisions. This threat is particularly notable in key applications such as disaster management, ecological tracking, and urban facilities monitoring, which have a critical requirement on the learning precision and EdgeIoT's reliability.

In this paper, a new manipulation model attack is proposed to compromise federated learning accuracy within EdgeIoT. Our approach employs an adversarial variational graph auto-encoder (AV-GAE) to generate malicious local model updates by exploring the features of benign local and global models. The attacker can eavesdrop on the local model updates uploaded from benign devices to the server. 

AV-GAE exhibits a remarkable ability to detect and analyze intricate patterns and structural relationships within graph-structured data, making it a powerful tool for federated learning-enabled EdgeIoT.
It excels at compressing graph data into a manageable, lower-dimensional space while preserving essential topological features. The attacker strategically alters the structure of the graph, maintaining the essential features of the models. This manipulation is specifically designed to inflate the training loss while disrupting the learning efficiency and degrading the performance of the global model. This altered graph structure is used to create harmful local models that align with the benign models' data characteristics. As a result, these manipulated local models can significantly disrupt the global model's integrity while remaining consistent with benign models, making detection of the AV-GAE-based attack particularly challenging.

This paper makes several important contributions:

\begin{itemize}
\item Introduction of a new model manipulation attack: We present a novel cyberattack designed to create data-independent manipulated model updates using a malicious IoT device. This method aims to reduce the accuracy of federated learning in EdgeIoT by altering correlations in benign model updates while maintaining the original data features.

\item Exploration of an AV-GAE attack: We examine a new attack model based on AV-GAE. This model manipulation attack is trained with the sub-gradient descent technique to subtly modify correlations within local models, ensuring that the manipulation remains undetectable.

\item Implementation and evaluation: Using CIFAR-10 and FashionMNIST datasets, our results reveal that the AV-GAE attack significantly disrupts federated learning performance in EdgeIoT. The performance demonstrates a marked drop of training accuracy, fluctuating between 50\% and 70\%, underscoring the attack's effectiveness in impairing federated learning efficiency.
\end{itemize}

The structure of this paper is as follows: Section~\ref{sec_relatedwork} reviews existing research on adversarial attacks within EdgeIoT and federated learning. Section~\ref{sec_systems} discusses the system model for federated learning in EdgeIoT, including aspects such as device interactions, and communication channels. In Section~\ref{sec_AV-GAE}, we describe the design and methodology of our AV-GAE attack. Section~\ref{sec_evaluation} details the performance evaluation of our approach. Finally, Section~\ref{sec_cond} presents the conclusion of the paper and future research.

\section{Related Work}
\label{sec_relatedwork}
In this section, we review related research on adversarial attacks against EdgeIoT and federated learning.

The work in~\cite{lyu2022privacy} emphasized the categorization of threats in federated learning. Poisoning attacks can be classified into two categories according to their intent: untargeted and targeted. Untargeted attacks strive to weaken the performance of the system, whereas targeted attacks aim to alter the system's behavior to generate specific, incorrect results. In~\cite{xia2023poisoning}, a summary of attacking and defending mechanisms in federated learning-enabled EdgeIoT was presented, with the attacks classified according to their techniques and objectives. Defense strategies were divided into three key approaches: model analysis, which examines models for potential tampering; Byzantine aggregation, which uses reliable techniques to reduce the influence of malicious models; and verification methods, which improves the integrity and authenticity before the aggregation. 

A collusion-based model poisoning attack was described in~\cite{tan2023collusive}, where malicious participants collaborated to generate untargeted poisoned local models within specific distance constraints, thereby reducing the global model's convergence and accuracy. Zhou et al.~\cite{zhou2021deep} designed a model poisoning attack by injecting malicious neurons into the unused regions of a neural network, resulting in a malicious local model. These neurons assisted in carrying out the attack but had minimal impact on the main task, ensuring the performance of the global model remains largely unaffected. 

The authors in~\cite{li2024data} studied a data-agnostic model poisoning attack on federated learning by developing a graph auto-encoder-based framework. This attack operated without requiring access to the FL training data, while achieving effectiveness and remaining undetectable. The attacker adversarially reconstructed these graph correlations, aiming to maximize the training loss in FL, and generated malicious local models by leveraging the adversarial graph structure and the data characteristics of the benign models. 

In~\cite{guo2022adfl}, a defense model was developed to counteract model poisoning in EdgeIoT. This model included a proof generation mechanism that allowed users to present evidence verifying whether their inputs were manipulated. Furthermore, it introduced an aggregation scheme aimed at preserving the training accuracy of the global model.

Existing poisoning attacks in federated learning often fail to account for the intricate interconnections between different local model updates. This limitation allows poisoning defense mechanisms, which rely on measuring Euclidean distances between local models, to detect such attacks. In contrast, the AV-GAE attack proposed in this paper for EdgeIoT introduces a new data-independent approach to model manipulation. Rather than focusing on raw data, this attack focuses on altering the relationships between features in specific benign local models. Importantly, it preserves the integrity of the data features within these models, ensuring that the poisoned local models generated by the malicious device remain undetectable.

\section{Federated Learning with IoT Devices}
\label{sec_systems}
We first examine the federated learning training process in EdgeIoT. We also formulate the channel gain between the benign IoT device and the server.

We focus on a system comprising $N$ benign IoT devices and $H$ authorized malicious device. A benign device $n$ ($\in [1, N]$) possesses a local dataset containing $B_n$ samples. The dataset consists of input data $x_c^n$ collected by device $n$ for the $c$-th sample, and $y_c^n$ which is the output generated by the model on that device for the same sample, where $c$ ranges from 1 to $B_n$~\cite{zheng2023federated}. We define $f_c(w_n; x_c^n, y_c^n)$ as a training loss function at device $n$, measuring the training error given $x_c^n$ and $y_c^n$. The weight of the loss function in the federated learning model being trained is denoted by $w_n$. For instance, $f_c(w_n; x_c^n, y_c^n)$ can be formulated by a linear regression, i.e., $f_c(w_n; x_c^n, y_c^n) = \frac{1}{2} (w_n^{\rm T} {x_c^n} - y_c^n)^2$, or a logistic regression, as in $f_c(w_n; x_c^n, y_c^n) = y_c^n \log \Big(1 + \exp \big(-w_n^{\rm T} {x_c^n}\big)\Big) - (1- y_c^n) \log \Big(1-\frac{1}{1 + \exp \big(-w_n^{\rm T} {x_c^n} \big)}\Big)$. The notation $(\cdot)^{\rm T}$ denotes the transpose.

Given $B_n$, the local loss function of device $n$ in every communication round can be formulated as:
\begin{align}
F_n(w_n) = \frac{1}{B_n} \sum_{c=1}^{B_n} f_c(w_n; x_c^n, \!y_c^n) + \alpha \mathcal{R}(w_n),
\label{eq_lossFunc}
\end{align}
where $\alpha \in [0,1]$ is a scaling coefficient, and $\mathcal{R}(\cdot)$ represents the regularization function accounting for the training noise at the IoT device. 

Moreover, we define $\mathcal{P}_n = (\mathcal{X}_n, \mathcal{Y}_n, \mathcal{Z}_n)$ as locations of the device $n$, while the location of the server is denoted by $\mathcal{P}_{\rm server} = (\mathcal{X}_{\rm server}, \mathcal{Y}_{\rm server}, 0)$. The distance between the IoT device $n$ and the server is given by
\begin{align}
\mathcal{D}_n &= ||\mathcal{P}_n - \mathcal{P}_{\rm server}|| \nonumber \\
&= \sqrt{(\mathcal{X}_n - \mathcal{X}_{\rm server})^2 + (\mathcal{Y}_n - \mathcal{Y}_{\rm server})^2 + \mathcal{Z}_n^2}. 
\end{align}

The channel gain between device $n$ and the server is given as $\mathcal{H}_n$, and $\mathcal{H}_n = \frac{\mathcal{T}_0}{\mathcal{D}_n^2}$, where we use $\mathcal{T}_0$ to represent a transmit power basis for the IoT device when $\mathcal{D}_n = 1$ m. Let $\mathcal{T}_i$ denote the transmit power of device $n$. We can have $\Lambda_n = \frac{\mathcal{H}_n \mathcal{T}_i}{\lambda_0^2}$, which presents the signal-to-noise ratio (SNR) of the channel between device $n$ and the server, where the noise level of the channel is $\lambda_0^2$.

\section{Proposed AV-GAE-based Model Manipulation Attacks}
\label{sec_AV-GAE}
In this section, we explore a threat scenario where a malicious IoT device intentionally crafts and sends altered local model updates. The adversary's goal is to gradually compromise the integrity of the global model by injecting these manipulated updates during the training process.

\subsection{Attacker in Federated Learning}
An attack model can be applied to the model manipulation attack in federated learning, where a malicious IoT device carries out the AV-GAE attack by intercepting benign local model updates. A manipulated local model update is generated at the malicious device and uploaded to the server. The manipulated update is specifically crafted to manipulate benign local models as well as global models in a detrimental way, reducing the model's accuracy and corrupting the benign local models. In federated learning-enabled EdgeIoT systems over wireless communications, model manipulation attacks are particularly concerning due to the broadcast nature of radio communication, which makes interception easier~\cite{li2025novel,li2022deep}. 

The malicious device is either a manipulated benign IoT device or an attacker, aiming to minimize the training accuracy of federated learning~\cite{yazdinejad2024robust}. In this attack scenario, the malicious IoT device methodically fabricates and uploads altered local models. This continuous injection of manipulated updates leads to the gradual degradation of the global model's integrity, which is represented by $\omega_g$. This, in turn, negatively impacts the benign local models, represented as $\omega_n$. In particular, $\omega_a$ denotes the manipulated model updates generated at the malicious IoT device (or attacker).

Without being aware of the malicious intent of the compromised IoT device, the server aggregates local model updates from all devices, including both benign and manipulated updates. This aggregation inadvertently results in the formation of a compromised global model, denoted as $\omega_g^a$. The total data samples can be given as $B = \sum_{n=1}^{N} B_n + B^a$, where $B^a$ represents the reported samples from the attacker. As a result, the manipulated global model can be expressed:

\begin{align}
\omega_g^a =  \sum_{n=1}^N  \frac{B_n}{B}\,\omega_n + \frac{B^a}{B}\omega_a,
\label{eq_glbAttacks}
\end{align}

\subsection{AV-GAE-based Model Manipulation Attack}
Based on the loss function of $F_i(\omega_n)$ and $\omega_g^a$, the manipulated model updates are crafted to maximize the loss function at the server, i.e., $\max{ F(\omega_g^a) }$. Additionally, we impose a constraint ensuring that the Euclidean distance between the malicious model $\omega_a$ and each benign model $\omega_n$ (for all $n \in \{1, \cdots, N\}$) remains below a predetermined threshold $D_{\rm thresh}$~\cite{chen2024exploring}. This limitation is crucial to make the malicious model appear similar to the benign ones, thereby avoiding detection by defense mechanisms that monitor for significant deviations in model updates~\cite{zheng2024detecting}.

\begin{figure}[htb]
\centering
\includegraphics[width=3.4in]{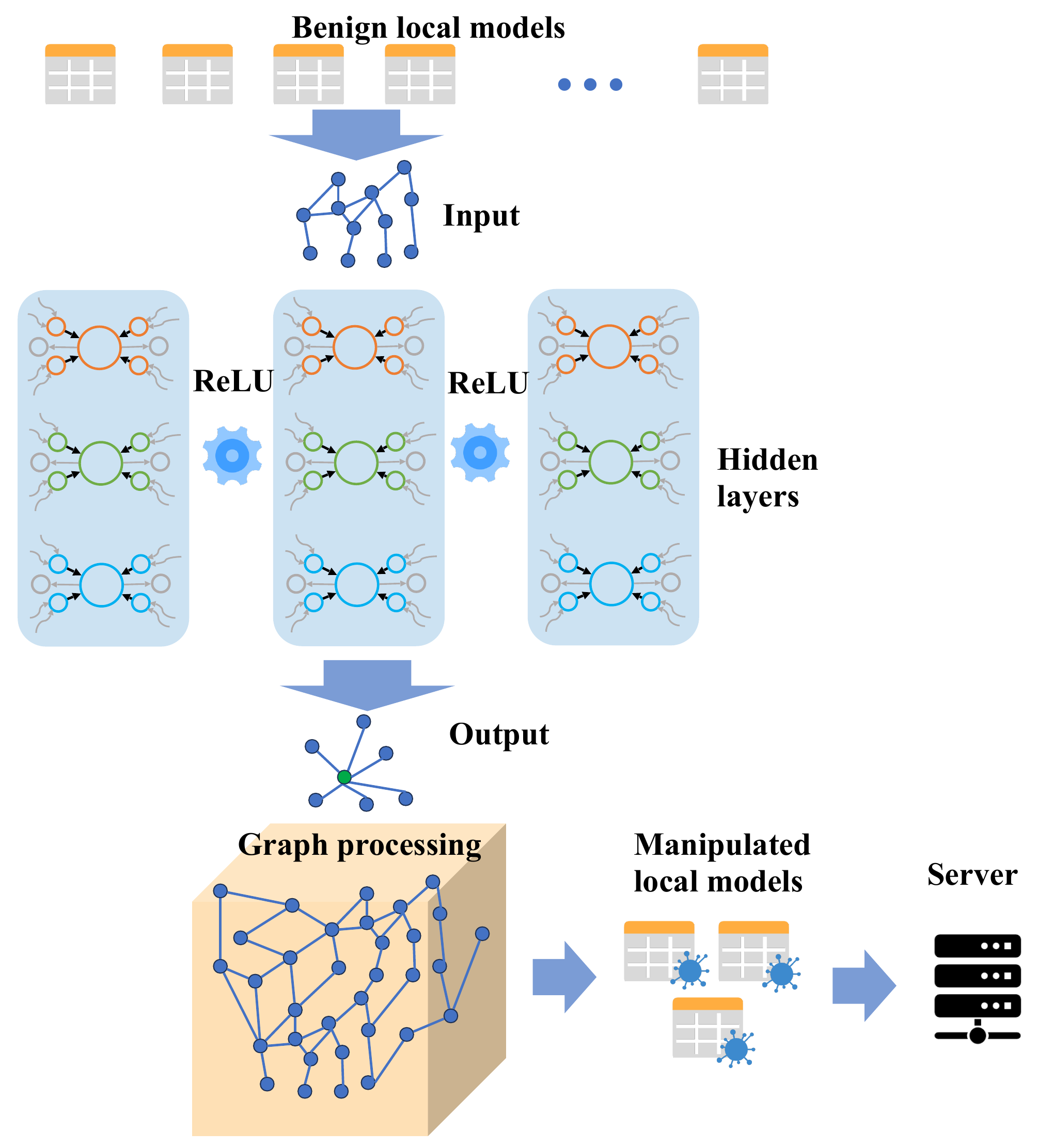}
\caption{Generating manipulated local models from the AV-GAE-based attack. }
\label{fig_AV-GAE}
\end{figure}

The AV-GAE attack seeks to solve $\omega_a^\star = \arg \max\{ F(\omega_g^a) \}$ by carefully preserving and manipulating the correlations between the generated malicious models and the benign ones, hindering the convergence of federated learning. Fig.~\ref{fig_AV-GAE} demonstrates the generation of manipulated local models from the AV-GAE-based attack, where the model parameters of benign IoT devices are decomposed into two components: a graph captures the correlations or similarities among the benign models~\cite{yin2024anomaly}, and data features represent benign local models. AV-GAE reconstructs the graph in a manipulative manner, where the manipulated models are generated by combining the reconstructed graph with original data features from benign models.

The AV-GAE attack allows the attackers to design the manipulated model update $\omega_a$ without the data knowledge of benign local models. In Fig.~\ref{fig_AV-GAE}, the graph $G$ is formulated according to the local models from the benign devices in EdgeIoT. Particularly, $V$ represents the vertice, $E$ presents the edge, and $Q$ denotes the features being involved in the graph. Moreover, $Q$ is defined as $Q = [\omega_1, \cdots, \omega_n, \omega_a]$, which includes the local models from the benign devices and the attacker. The proposed AV-GAE architecture comprises inputs, outputs, and $L$ number of hidden layers. In the $l$-th layer, $\eta_V^l$ is a learnable weight vector associated with the edges of the vertices in $V$. The hidden state of the vertices $V$ at the $l$-th layer is
\begin{align}
&\kappa^l_V = \theta^l\Big(\kappa^{l-1}_V \oplus {\cal A}^l \Big( \{\kappa^{l-1}_{V^\prime, E} : (V, V^\prime) \in E^l\}_{E^l \in \mathbb{R}^{E}}\Big); \eta_V^l\Big),
\label{eq_features}
\end{align}
where $\oplus$ indicates a summation operation over embeddings, and $\theta^l(\cdot)$ represents a nonlinear activation function. Examples of such activation functions include $\tanh(\cdot)$ (the hyperbolic tangent function) and ${\rm ReLU}(\cdot)$ (the Rectified Linear Unit)~\cite{parhi2020role}. $\kappa_V$, $\kappa_E$, and $\kappa_{V^\prime}$ refer to the representation of $V$, $E$, and the neighboring vertice $V^\prime$, respectively. $E^l$ provides the edge in the $l$-th layer, and $\mathbb{R}^{E}$ gives the number of hidden states. ${\cal A}^l(\cdot)$ refers to an aggregation function, which gathers neighbor's knowledge of multiple correlations into a single vector. In addition, the initialized value of $\kappa^l_V$ is $\kappa^0_V = V$.

According to~\eqref{eq_features}, $\eta_V^l$ can be optimized by minimizing the graph generation loss, denoted as $\delta^l_G$, which is
\begin{align}
\delta^l_G = \sum_{V \in V_G} -\log \Big( \theta^l(\Psi(\kappa^l_V)) \Big),
\label{eq_gnn_loss}
\end{align}
where $\Psi$ represents a multilayer perceptron (MLP). At layer $l$, the function $\Psi$ can be formulated as a hyperbolic tangent activation function $\theta^l(\cdot)$. It takes the node embeddings generated from the $l-1$-th layer as an input. The scalar output produced by $\Psi$ is then passed through another activation function denoted as $\theta^l(\cdot)$.

\begin{algorithm}[t]
\caption{Algorithm of the AV-GAE-based model manipulation attack}
\label{alg_AV-GAE}
\begin{algorithmic}[1] \STATE{\textbf{1. Initializing}: $N$, $H$, $B_n$, $D_{\rm thresh}$, $\omega_g^a$, and $\omega_a$.} \\ {\textbf{Learning Process}}: \FOR{Communication rounds $m$ = 1, 2, 3, ...} \FOR{Learning iterations $t_m$ = 1, 2, 3, ...} \STATE{For each benign device $n$, train local models using its dataset samples $B_n$ as per~\eqref{eq_lossFunc} $\to \omega_n(t_m)$.} \ENDFOR \STATE{Benign devices upload their local model updates $\omega_n(m)$, $n=1,\cdots, N$, to the server. The malicious device overhears the updates from neighboring devices.} \\ 
{\textbf{Executing the AV-GAE Attack}}: \STATE{The malicious device $h (\in \{1, \cdots, H\})$ formulates the graph $G$ to generate its manipulated updates $\omega_a(m)$:} \FOR{$V \in V_G$} \FOR{Layer $l$ = 1 to $L$} \STATE{For every $V$, compute the hidden state using~\eqref{eq_features} $\to \kappa^l V(m)$.} \STATE{Compute the graph generation loss $\delta^l_G(m)$ using~\eqref{eq_gnn_loss}.} \STATE{Optimize $\eta_V^l$ to minimize the graph loss $\mathcal{L}^l_G$.} \ENDFOR \ENDFOR \STATE{Determine the optimal manipulated model update $\omega_a^\star(m) = \arg \max{ F(\omega_g^a(m)) }$.} \STATE{The attacker transmits its manipulated model updates $\omega_a(m)$ to the server.} \STATE{Benign and manipulated local models are aggregated by the server, using~\eqref{eq_glbAttacks} to produce the global model $\omega_g^a(m)$, which is broadcast to all devices.} \STATE{{$\omega_n(m) \leftarrow \omega_g^a(m),\forall i$}.} \ENDFOR \end{algorithmic}
\end{algorithm}

\subsection{Algorithm Design of The AV-GAE-based Model Manipulation Attack}
Algorithm~\ref{alg_AV-GAE} illustrates how the malicious IoT device executes the proposed AV-GAE attack in EdgeIoT to generate manipulated local model updates, which are then transmitted to the server for federated learning. The number of attackers is denoted as $H$.

\begin{figure}[htb]
\centering
\includegraphics[width=3.6in]{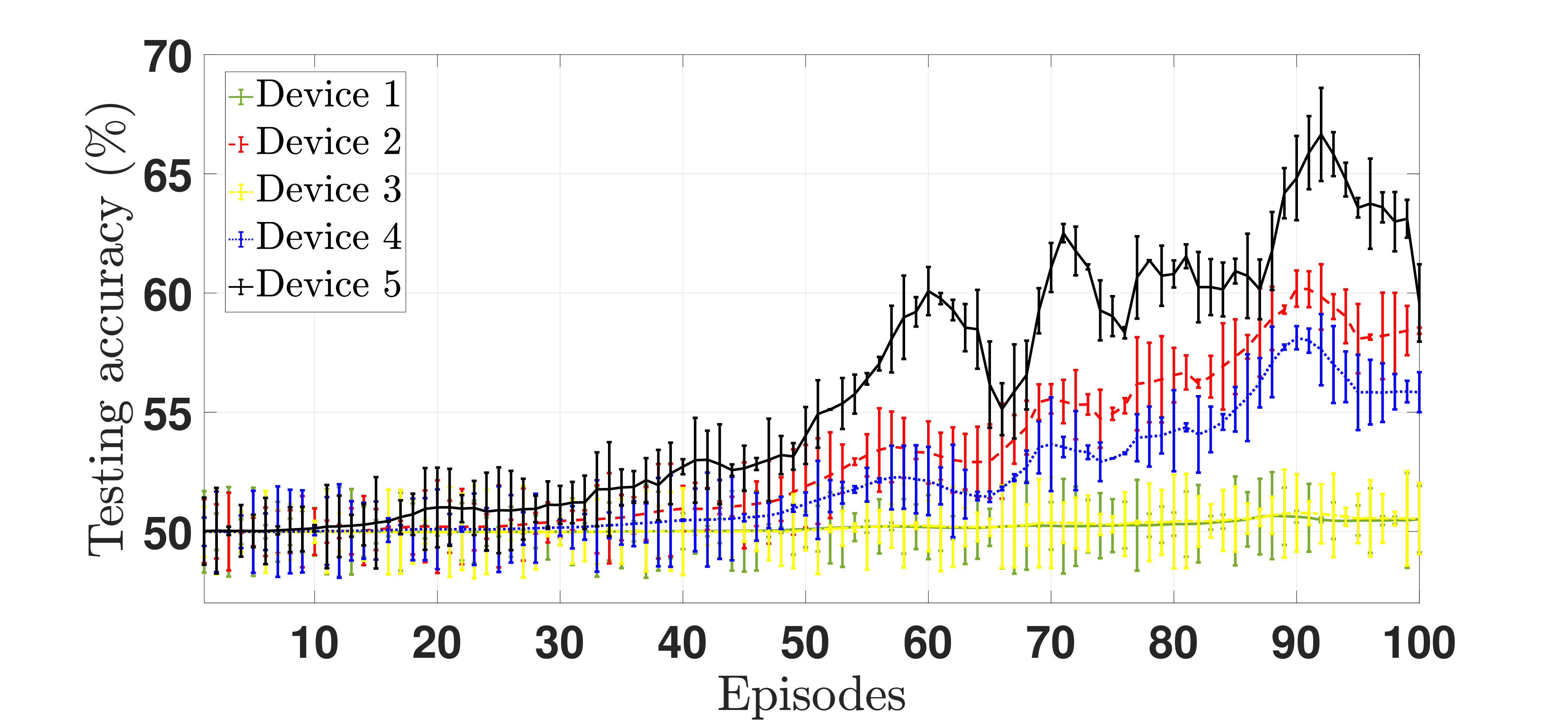}
\caption{The testing accuracy with CIFAR-10.}
\label{fig_accuracy_cifar10}
\end{figure}

\begin{figure}[htb]
\centering
\includegraphics[width=3.6in]{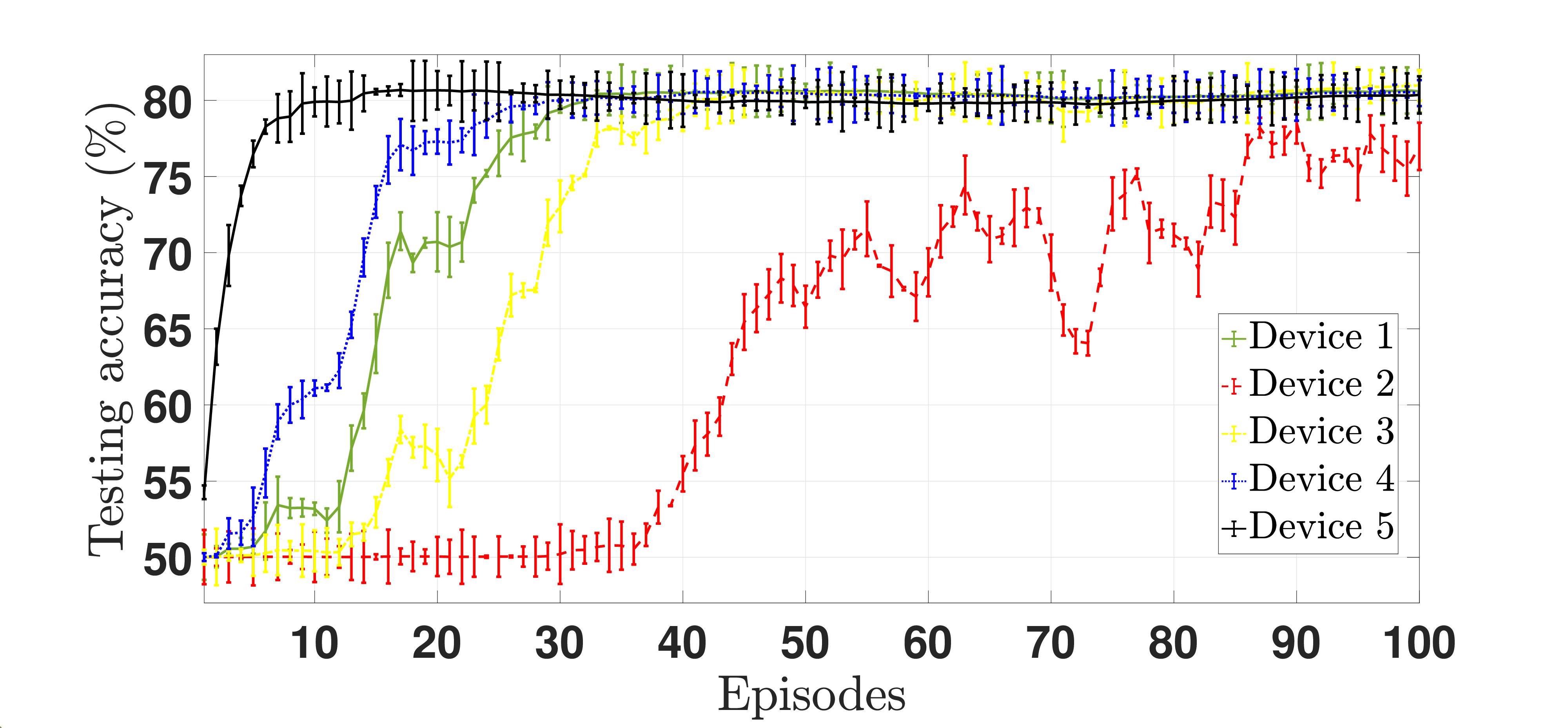}
\caption{The testing accuracy with FashionMNIST.}
\label{fig_accuracy_fashionmnist}
\end{figure} 

Initially, the AV-GAE-based model manipulation attack is set up at the malicious IoT device with parameters including the graph structure $G$, representing the relationships between devices, and various other initial values, such as, the model weights and dataset sizes. During each communication round, benign devices locally train their models based on their datasets and send these updates to the server. Meanwhile, the malicious device overhears the local model updates of neighboring devices. The malicious device then employs the AV-GAE attack by decomposing the local models into graph structures and feature embeddings, using the presented multi-layered architecture to iteratively compute hidden states and minimize graph generation loss. This allows the malicious device to strategically manipulate the model correlations and generate an optimized manipulated model update, $\omega_a$, designed to maximize the global model’s training loss. The server, unaware of the attack, aggregates both benign and manipulated updates to create a compromised global model, which is then broadcast back to all devices, including the benign ones. These devices then update their local models based on the compromised global model, propagating the attack throughout the federated learning process.

\section{Performance Analysis}
\label{sec_evaluation}
This section presents the accuracy of federated learning in EdgeIoT as well as the Euclidean distances, based on the CIFAR-10 and FashionMNIST datasets.

\begin{figure}[htb]
\centering
\includegraphics[width=3.6in]{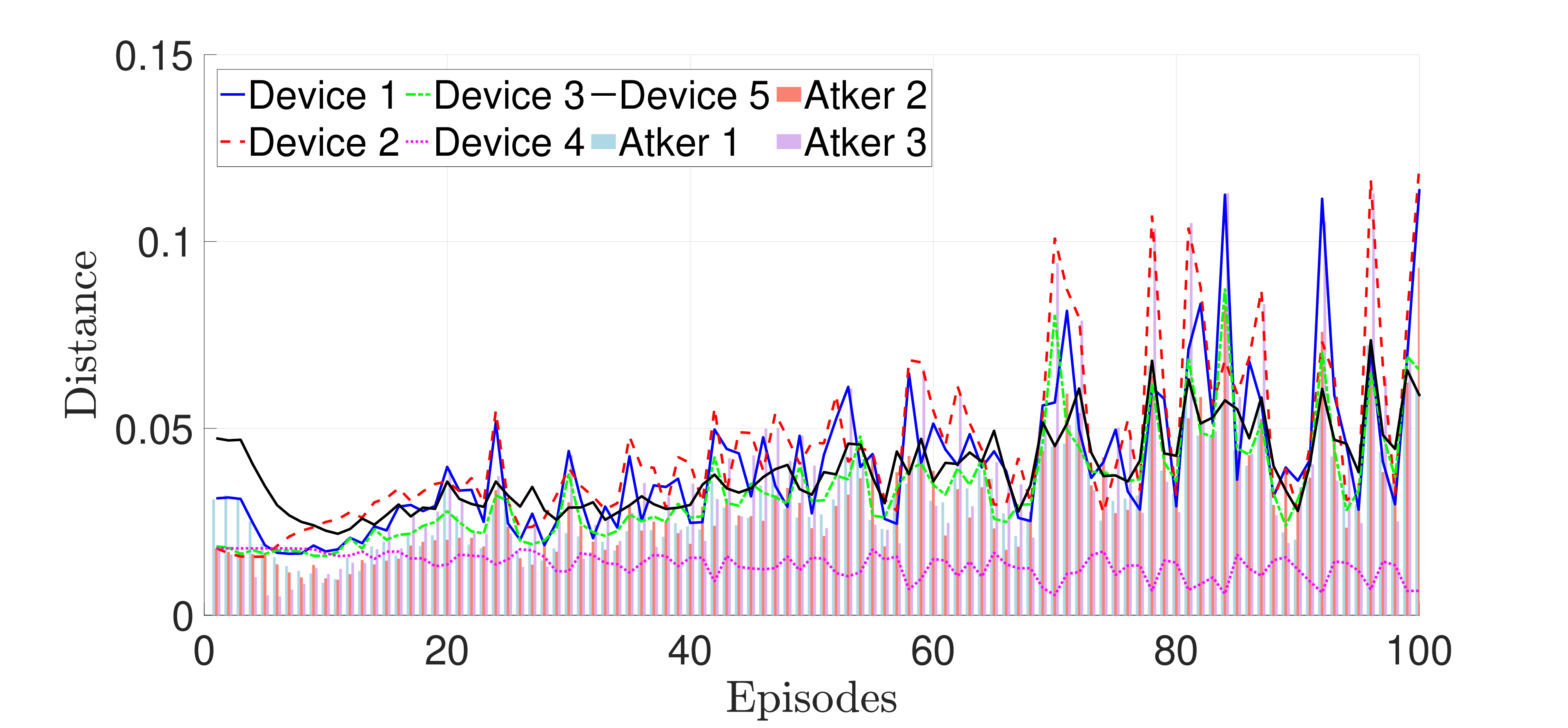}
\caption{The Euclidean distances under the AV-GAE attack, where three malicious devices are denoted by ``Atker 1'', ``Atker 2'', and ``Atker 3'', respectively.}
\label{fig_distance_avgae}
\end{figure}

\begin{figure}[htb]
\centering
\includegraphics[width=3.6in]{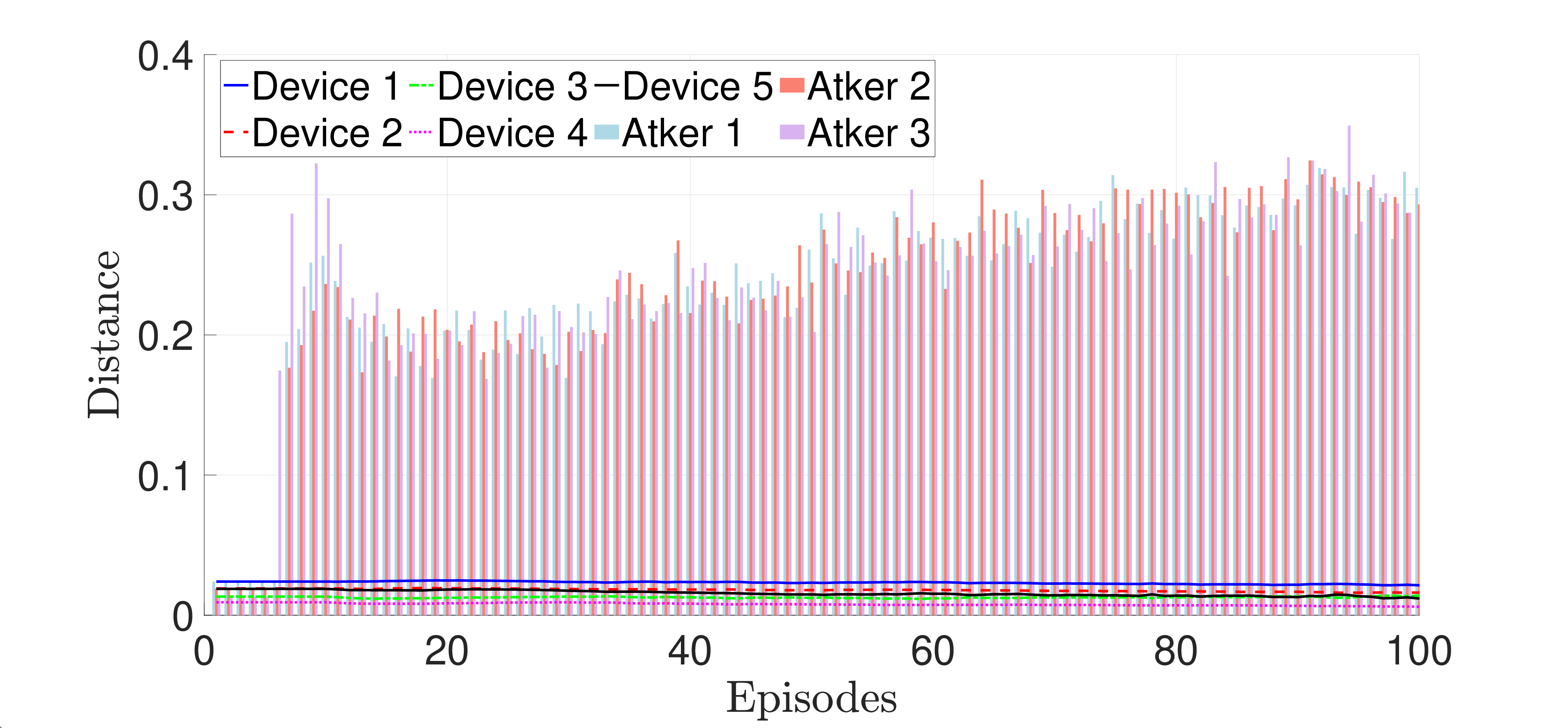}
\caption{The Euclidean distances under the existing model poisoning attack, where three malicious devices are denoted by ``Atker 1'', ``Atker 2'', and ``Atker 3'', respectively.}
\label{fig_distance_mp}
\end{figure}

Given 100 communication rounds, five benign devices and two malicious devices, Figs.~\ref{fig_accuracy_cifar10} and~\ref{fig_accuracy_fashionmnist} illustrate the testing accuracy of the local model across 100 communication rounds under the proposed AV-GAE attack. The results are shown for two datasets: CIFAR-10 and FashionMNIST. The plots provide insights into how the AV-GAE attack affects the testing accuracy of the local models during the training process, highlighting variations in performance across different datasets. Specifically, the AV-GAE attack causes the testing accuracy on the CIFAR-10 and FashionMNIST datasets to fluctuate between 50\% and 70\%, and 50\% and 80\%, respectively. This fluctuation occurs because the manipulated models disrupt the training convergence of the federated learning process. The newly proposed AV-GAE attack reconstructs the adversarial adjacency matrix based on the unique features of each IoT device. As a result, the malicious IoT device falsifies local models, effectively maximizing the federated learning loss and hindering model accuracy and stability.

To assess the stealthiness of the proposed AV-GAE attack, we analyze the distance between the local and global models using the CIFAR-10 dataset, as shown in Figs.~\ref{fig_distance_avgae} and~\ref{fig_distance_mp}. In this evaluation, the number of benign IoT devices is set to $N = 5$, and the number of malicious devices is $H = 3$, i.e., ``Atker 1'', ``Atker 2'', and ``Atker 3''. This study helps determine how closely the manipulated local models resemble the global model, providing insights into the AV-GAE attack's ability to remain undetected while influencing the federated learning process. 

As shown in Fig.~\ref{fig_distance_avgae}, the Euclidean distances between the malicious local models generated by the proposed AV-GAE attack and the corresponding global models are smaller than those of the benign local models. This reduced distance makes it challenging for the server to detect and defend against the manipulated model updates. 

In contrast, Fig.~\ref{fig_distance_mp} illustrates that the existing model poisoning attacks result in significantly larger distances between the malicious local models and the global model, making them easier to identify. This demonstrates a key advantage of the AV-GAE-based attack, which is its ability to generate manipulated local models that closely mimic the feature correlations between the benign local models and the global model, making the differences between manipulated and benign models indistinguishable.

\section{Conclusions and Future Research}
\label{sec_cond}
This paper examined the impact of model accuracy manipulation attacks on federated learning-enabled EdgeIoT, where machine learning models were trained locally on devices and aggregated by a server to refine a global model. We proposed a novel data-independent model accuracy manipulation attack that does not rely on the training data from EdgeIoT devices. This attack utilized an AV-GAE to generate malicious local model updates by analyzing benign local models observed during communication. The AV-GAE attack demonstrated proficiency in identifying and interpreting the structural relationships within the graph representations of these benign models, as well as the data features that underpin them. By reconstructing graph structures, the attacker could create manipulated local model updates that adversely affect the global model.

Future research into employing AV-GAE for manipulating federated learning-enabled EdgeIoT holds promise for advancing offensive and defensive strategies. The AV-GAE's ability to model intricate relationships and dependencies in data positions is a powerful tool for developing model accuracy manipulation attacks designed to the EdgeIoT. Adversarial manipulation, embedded within GAE-based representations, can more effectively disrupt federated learning compared to traditional poisoning methods. On the defensive front, there is growing interest in creating advanced strategies for the detection of AV-GAE-based attacks, which examines graph features for anomalies or manipulated models. New solutions have to be developed for enhancing federated learning security, especially in critical applications related to safety.

\section*{Acknowledgements}
This work was supported by the CISTER Research Unit (UIDP/UIDB/04234/2020) and project ADANET (PTDC/EEICOM/3362/2021), financed by National Funds through FCT/MCTES (Portuguese Foundation for Science and Technology).

\ifCLASSOPTIONcaptionsoff
  \newpage
\fi

\bibliographystyle{IEEEtran}
\bibliography{bibAVGAE}  

\end{document}